\documentclass[10pt,twocolumn,letterpaper]{article}

\usepackage{cvpr}
\usepackage{times}
\usepackage{epsfig}
\usepackage{graphicx}
\usepackage{amsmath}
\usepackage{amssymb}
\usepackage{booktabs}
\usepackage{multirow}
\usepackage{xcolor}
\usepackage{authblk}

\usepackage[pagebackref=true,breaklinks=true,letterpaper=true,colorlinks,bookmarks=false]{hyperref}

\cvprfinalcopy 

\def\cvprPaperID{71} 

\ifcvprfinal\fi

\begin{document}

\title{Unsupervised Real Image Super-Resolution via Generative Variational AutoEncoder}

\author[1,2]{Zhi-Song Liu\thanks{Most work was done in The Hong Kong Polytechnic University.}}
\author[1]{Wan-Chi Siu}
\author[1]{Li-Wen Wang}
\author[1]{Chu-Tak Li}
\author[2]{Marie-Paule Cani}
\author[1]{Yui-Lam Chan}
\affil[1]{The Hong Kong Polytechnic University} 
\affil[2]{LIX, École polytechnique}

\maketitle
\thispagestyle{empty}

\begin{abstract}
	Benefited from the deep learning, image Super-Resolution has been one of the most developing research fields in computer vision. Depending upon whether using a discriminator or not, a deep convolutional neural network can provide an image with high fidelity or better perceptual quality. Due to the lack of ground truth images in real life, people prefer a photo-realistic image with low fidelity to a blurry image with high fidelity. In this paper, we revisit the classic example based image super-resolution approaches and come up with a novel generative model for perceptual image super-resolution. Given that real images contain various noise and artifacts, we propose a joint image denoising and super-resolution model via  Variational AutoEncoder. We come up with a conditional variational autoencoder to encode the reference for dense feature vector which can then be transferred to the decoder for target image denoising. With the aid of the discriminator, an additional overhead of super-resolution subnetwork is attached to super-resolve the denoised image with photo-realistic visual quality. We participated the NTIRE2020 Real Image Super-Resolution Challenge ~\cite{NTIRE2020RWSRchallenge} . Experimental results show that by using the proposed approach, we can obtain enlarged images with clean and pleasant features compared to other supervised methods. We also compared our approach with state-of-the-art methods on various datasets to demonstrate the efficiency of our proposed unsupervised super-resolution model.
\end{abstract}

\section{Introduction}
\begin{figure}[t]
	\vskip 0.01in
	\begin{center}
		\centerline{\includegraphics[width=0.9\columnwidth]{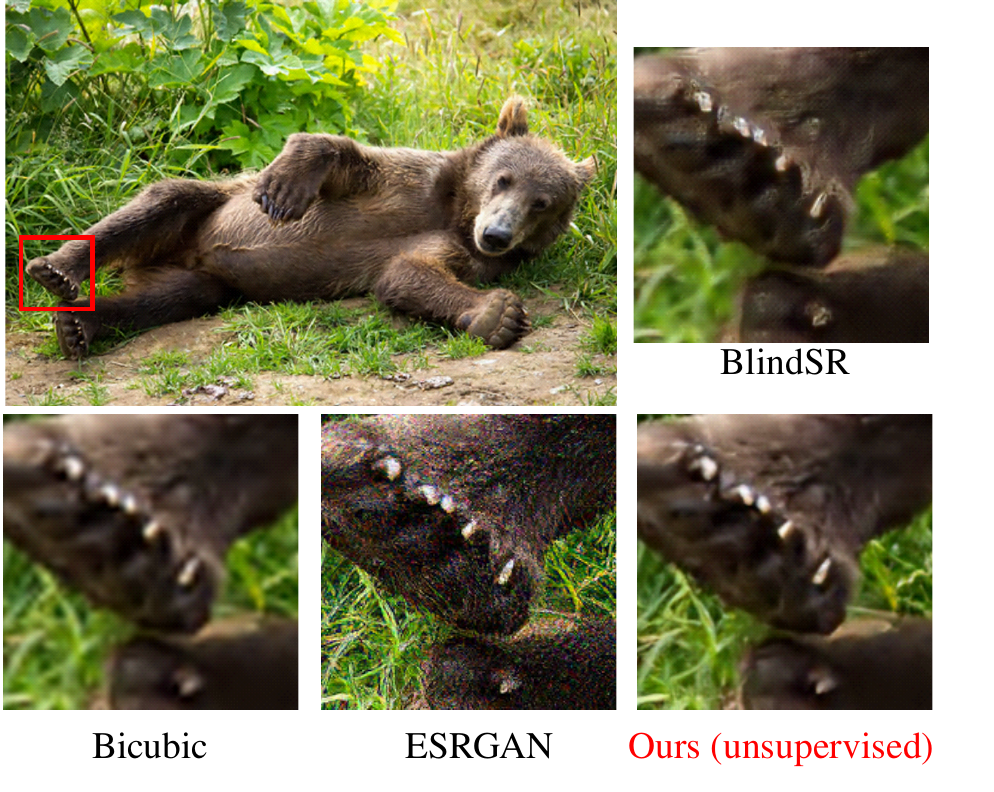}}
		\caption{Visualization of SR result on NTIRE2020 validation image. The red one is our proposed dSRVAE approach.}
		\label{Figure 1}
	\end{center}
	\vskip -0.3in
\end{figure}
Example based image Super-Resolution (SR) is a classic supervised learning approach that has inspired many SR works. The concept is based on that same patterns are likely repetitive across the whole images. In order to fill out the missing pixels accurately, researchers have proposed many approaches to model the image patterns for prediction. 

Pixel based interpolation is one of the earliest learning based SR approaches. It models a group of pixels by assuming the geometry duality across the neighbourhoods. The problem is that individual pixel contains very little information. The assumption can only hold in a small region. To better grasp the pattern, patch based SR approaches~\cite{KSVD,NCSR,A+,SRRMF,CRFSR,ISCAS17,ICIP18} were proposed dominanting the research approaches for a long time. Similarly, researchers use patches rather than pixels based on the piece-wise linearity. The complete image can be divided into many patches and each patch can be modelled by a simple linear regression. Similar patches not only can be found in the image itself, but also from external images. Hence there are substantial research works investigating internal or external based image SR. In order to improve the SR quality, more data are exploited for patch clustering and regression but it can quickly become cumbersome and over complex. Convolutional Neural Network (CNN) works better than most machine learning approaches because it can digest huge amount of data to learn different filters for feature extraction via backpropagation. Many CNN based SR approaches~\cite{SRCNN,VDSR,LapSRN,EDSR,RCAN,DBPN,HBPN,ABPN,FrequencySF,RefSR,ZSSR,BlindSR,SRGAN,ESRGAN,NTIRE2019,NTIRE2020RWSRchallenge,AIM2019RWSRchallenge,lugmayrICCVW2019} have successfully boosted up the image super-resolution performance in both computation and quality. 

Most algorithms of the pixel and patch based approaches rely on supervised learning. They require paired low-resolution (LR) and ground truth high-resolution (HR) images for building the reconstruction mapping. In order to mimic the real images, the most common process is to use HR images to simulate LR images by a spatial domain down-sampling (Bicubic process) or transform domain down-sampling (DCT and Wavelet process). However, this kind of simulation still simplifies the real situation where real images could also be degraded by different noises or photo editing. Better simulation has been proposed to use cameras to capture LR and HR images with different focal lens and then align the pixels by image registration~\cite{NTIRE2019}. 

Though researchers came up with different simulations to model the down-sampling process, it still targets on one specific applications. Real-world super-resolution is far more complicated. As investigated in~\cite{AIM2019RWSRchallenge,NTIRE2020RWSRchallenge}, there is no available ground-truth LR-HR image pairs. Most supervised image SR approaches have the overfitting problem. As shown in Figure~\ref{Figure 1}, once the down-sampling is different from the assumption, supervised approaches fail while our proposed method can generate robust and good results. Instead of learning the reconstruction in supervised manner, in this work, we propose a novel unsupervised real image denoising and Super-Resolution approach via Variational AutoEncoder (dSRVAE). We add denoising task to super-resolution because real images usually contain various types of noises and degradations. With the lack of targeted HR images, pursuing lower pixel distortion may lose its meanings. According to Generative Adversarial Network (GAN) based SR approaches~\cite{BlindSR,SRGAN,ESRGAN,CycleGAN}, a discriminator can constrain the network to generate photo-realistic quality for sacrificing image distortion. Based on this observation, the proposed network is made of two parts: Denoising AutoEncoder (DAE) and Super-Resolution Sub-Network (SRSN) with an attached discriminator. Contrast to previous works, we claim the following points:
\begin{enumerate}
	\item To the best of our knowledge, this is the first work on joint real image denoising and super-resolution via unsupervised learning.
	\item This is also the first work on combining Variational AutoEncoder and Generative Adversarial Network for image super-resolution.           
	\item To stabilize the adversarial training, we propose a simple cycle training strategy to force the network to balance the reference and super-resolved images.
\end{enumerate}

\section{Related Work}
In this section, we give a brief review of previous works related to our proposed method. We focus on perceptual image super-resolution, hence omitting a main body of works on generative approaches for image super-resolution. We also introduce the related unsupervised learning for image super-resolution, like blind image SR. Interested readers may refer to these literatures for more details.

\subsection{Perceptual Image Super-Resolution}
In the past few years, it has been widely observed that there is a tradeoff between distortion and perception. SR approaches on reducing the pixel distortion tend to generate over-smooth results. For practical applications, with the absence of ground truth images, researchers are more attracted to the images with distinct textures (even fake ones). Generative Adversarial Network~\cite{BlindSR,SRGAN,ESRGAN} adopted by many SR approaches has the ability to provide photo-realistic images. The basic idea is to train a generator well enough that the discriminator cannot distinguish the SR images from HR images. Additional pre-trained deep networks are usually used to measure the key feature losses. SRGAN~\cite{SRGAN} is the first work using GAN for perceptual image SR. It uses VGG feature maps to allow visually pleasant image generation. ESRGAN~\cite{ESRGAN} further improves the visual quality by replacing the standard discriminator to relativistic discriminator that sharpens the textures and edges. In terms of the evaluation of perceptual quality, some works~\cite{PIRM,LPIPS} were proposed to measure the visual quality by handcrafted or automatic criteria. For instance, Learned Perceptual Image Patch Similarity (LPIPS) metric is the one using various deep network activations to score the image visual quality. 
\begin{figure*}[h]
	\vskip 0.01in
	\begin{center}
		\centerline{\includegraphics[width=0.65\textwidth]{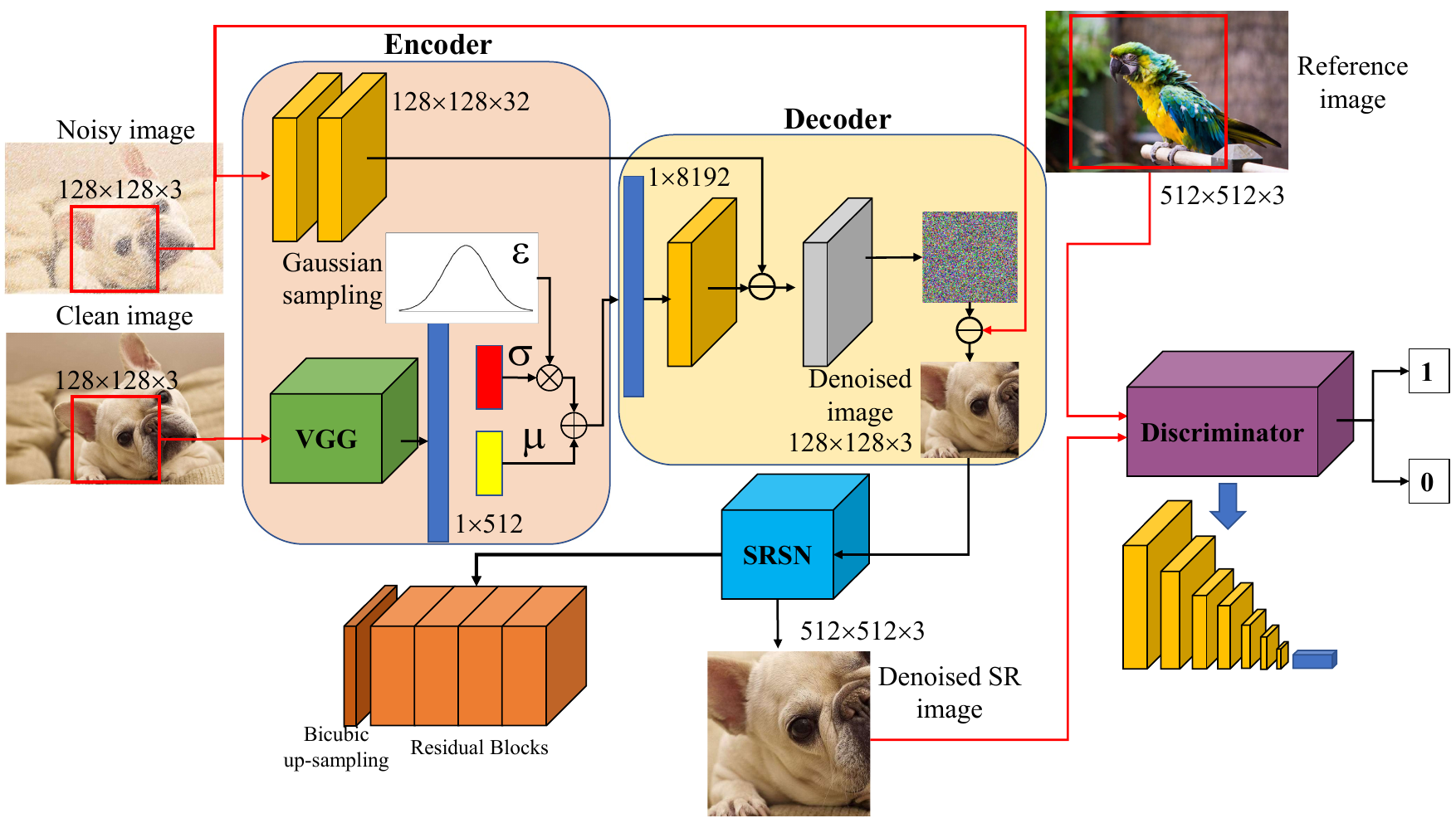}}
		\caption{Complete structure of the proposed dSRVAE model. It includes Denoising AutoEnocder (DAE) and Super-Resolution Sub-Network (SRSN). The discriminator is attached for photo-realistic SR generation.}
		\label{Figure 2}
	\end{center}
	\vskip -0.3in
\end{figure*}

\subsection{Real-World Super-Resolution}
Given the fact that a real image contains more complicated noise and artifacts, real world super-resolution is proposed to resolve the problem. There are two features of ``real-world'' super-resolution: 1) online training and testing and 2) Estimating degradation factor using prior information. One of the representative work is ZSSR~\cite{ZSSR}. It use the low-resolution image itself to learn internal statistics for super-resolution. No prior information is required for training. It can be considered as the first CNN based unsupervised image SR approach. On the other hand, with the huge learning capacity of deep neural network, we can assume degradation factors in low-resolution image generation, like adding different noise levels, forming blur kernels with some combinations of scale factors, etc., and then combine various of these factors for a general image super-resolution. For example, we can have joint demosaicing and super-resolution~\cite{Demosaic_SR_1,Demosaic_SR_2}, joint denoising and super-resolution~\cite{Denoise_SR} and joint deblurring, denoising and super-resolution~\cite{BlindSR,ss,CycleGAN}. Considering that the real images are normally obtained by unknown or non-ideal process, it is cumbersome or even impossible to include all the degradation factors in the training phase. A better real world image SR should be learned from an unsupervised approach, where the ground truth images are not involved in training stage.

\section{The Proposed Method}
In the following section, we will give a detailed introduction of our proposed work. Let us formally define the real image SR. Mathematically, given a LR image $\mathbf{X}\in\mathbb{R}^{\mathit{m}\times\mathit{n}\times3}$ which may be down-sampled from an unknown HR image $\mathbf{Y}\in\mathbb{R}^{\mathit{\alpha m}\times\mathit{\alpha n}\times3}$, where ($\mathit{m}$, $\mathit{n}$) is the dimension of the image and $\alpha$ is the up-sampling factor. They are related by th following degradation model,
\begin{small}
\begin{equation}
\mathbf{X}=\mathbf{sKY}+\mu \tag{1}
\label{Equation 1}
\end{equation}
\end{small}\\
where $\mu$ is the additive noise, \textbf{s} is the down-sampling operator and \textbf{K} is the blur kernel. The goal of image SR is to resolve Equation~\ref{Equation 1} as a Maximum A Posterior (MAP) problem as follows,
\begin{small}
\begin{equation}
\mathbf{\hat{Y}}=\underset{\mathbf{Y}}{\arg\max}\, log \mathit{P}(\mathbf{X|Y})+log \mathit{P}(\mathbf{Y}) \tag{2}
\label{Equation 2}
\end{equation}
\end{small}\\
where $\mathbf{\hat{Y}}$ is the predicted SR image. log$\mathit{P}$($\mathbf{X}|\mathbf{Y}$) represents the log-likelihood of LR images given HR images and log$\mathit{P}$($\mathbf{Y}$) is the prior of HR images that is used for model optimization. Formally, we resolve the image SR problem as follows,
\begin{small}
\begin{equation}
\underset{\theta}{\min} \Arrowvert\mathbf{Y-\hat{Y}}\Arrowvert^\mathit{r}  \ \text{s.t.} \mathbf{\hat{Y}}=\underset{\mathbf{Y}}{\arg\min} \frac{1}{2}\Arrowvert\mathbf{X}-\mathbf{sKY}\Arrowvert^2+\lambda\Omega(\mathbf{Y}) \tag{3}
\label{Equation 3}
\end{equation}
\end{small}\\
where $\Arrowvert\ast\Arrowvert^\mathit{r}$ represents the $\mathit{r}$-th order estimation of pixel based distortion. The regularization term $\Omega(\mathbf{\mathbf{Y}})$ controls the complexity of the model. The noise pattern is omitted in Equation~\ref{Equation 3} on the assumption that the noise is independent from the signal and the residual between the estimation and the ground truth can be optimized by various linear or non-linear approaches. In the real world, the noise comes from the camera sensor or data compression and it is signal-dependent. Direct super-resolution usually fails to generate clean images. For practical application, a generalized super-resolution model is required to handle various degradations and distortions. It would be useful to firstly decouple the noise from the LR image and then performs super-resolution. Meanwhile, this disentanglement process can also be beneficial to real applications. As shown in Figure~\ref{Figure 2}, we propose a joint image denoising and Super-Resolution model by using generative Variational AutoEncoder (dSRVAE). It includes two parts: Denoising AutoEncoder (DAE) and Super-Resolution Sub-Network (SRSN). With the absence of target images, a discriminator is attached together with the autoencoder to encourage the SR images to pick up the desired visual pattern from the reference images.  The details of the structure will be discussed in the following parts.

\subsection{Denoising AutoEncoder (DAE)}
Mathematically, Conditional Variational AutoEncoder (VAE) can be formed as follows,
\begin{small}
\begin{equation}
P(\mathbf{Y|X})=\int P(\mathbf{Y|X},z)P(z|\mathbf{X})\, dz \tag{4}
\label{Equation 4}
\end{equation}
\end{small}\\
where vector z is sampled from the high-dimensional space $\mathbf{Z}$. VAE targets on learning the latent variable that describes the conditional distribution. We can use Bayesian rule to rewrite  Equation~\ref{Equation 4} as
\begin{small}
\begin{equation} \tag*{(5)}
\begin{matrix}
\begin{split}
logP_{\theta}(\mathbf{Y|X})&\ge\int log P(\mathbf{Y|X},z)log P(z|\mathbf{X})\, dz \\
&=E_{Q_{\theta}(z|X)}\left[log\frac{P_{\theta}(\mathbf{Y|X},z)P_{\theta}(z|\mathbf{X})}{Q_{\phi}(z|\mathbf{X,Y})}\right]
\end{split}
\end{matrix} 
\label{Equation 5}
\end{equation}
\end{small}\\
We design the network to learn parameters $\theta$ for maximizing the data log likelihood $P_{\theta}(\mathbf{Y|X})$. Equation~\ref{Equation 5} can be further rearranged as the following equation,
\begin{small}
\begin{equation} \tag*{(6)}
\begin{matrix}
\begin{split}
logP_{\theta}(\mathbf{Y|X})&=E_{Q_{\theta}(z|X)}\left[log\frac{P_{\theta}(\mathbf{Y|X},z)P_{\theta}(z|\mathbf{X})}{Q_{\phi}(z|\mathbf{X,Y})}\right] \\
&=E_{Q_{\phi}(z|X)}[logP_{\theta}(\mathbf{Y|X},z)] \\
&-KL[Q_{\phi}(z|\mathbf{X,Y})|P_{\theta}(z|\mathbf{X})]
\end{split}
\end{matrix} 
\label{Equation 6}
\end{equation}
\end{small}\\
where $KL[p|q]$ represents the KL divergence. Equation~\ref{Equation 6} can be interpreted in the way that the encoder is designed to learn a set of parameters $\phi$ to approximate posterior $Q_{\phi}(z|\mathbf{X,Y})$, while the decoder learns parameters $\theta$ to represent the likelihood $P_\theta(\mathbf{Y|X},z)$. We can adopt the KL divergence to represent the divergence between predicted distributions $Q_{\phi}(z|\mathbf{X,Y})$ and $P_\theta(\mathbf{Y|X},z)$. In order to compute the gradients for backpropagation, the ``reparameterization trick''~\cite{VAE} is used to randomly sample from $Q_{\phi}(z|\mathbf{X,Y})$ and then compute latent variable as $z=\mu(\mathbf{X,Y})+\varepsilon\ast\sigma^{0.5}(\mathbf{X,Y})$. 

To utilize variational autoencoder for image denoising, the posterior needs to be modified from $P(\mathbf{Y|X})$ to $P(\mathbf{T|X})$, where $\mathbf{T}$ is the target clean image. The encoder compresses the clean image to learn the latent variables. Then the decoder learns to extract the noise from the noisy image and the sampled vector $\mathit{z}$. Results in~\cite{SRCNN,ESRGAN} show that VGG19 network~\cite{VGG} is a good feature extractor for image processing, we discard the fully connected layers and only use the rest of convolution layers as the encoder to extract feature maps from the clean image. Let us mathematically define the training process of DAE as follows.
\begin{small}
\begin{equation} \tag*{(7)}
\begin{matrix}
\begin{split}
\frac{1}{N}&\sum_n^N logP_{\theta}(\mathbf{T_n|X_n}) \\
&=\frac{1}{N}\sum_n^N E_{Q_{\theta}(z|X_n)}\left[log P_{\theta}(\mathbf{T|X},z=\mu+\varepsilon\ast\sigma^{0.5})\right] \\
&-KL[Q_{\phi}(z|\mathbf{X_n,T_n})|P_\theta(z|\mathbf{X_n})]
\end{split}
\end{matrix} 
\label{Equation 7}
\end{equation}
\end{small}\\
where \textit{N} is the batch number. The output of the decoder is the estimation of the noise pattern. By subtracting it from the real LR image, we can obtain the clean image for the following super-resolution process. During the testing, the encoder can be discarded and only the decoder is needed for image denoising.
\subsection{Super-Resolution Sub-Network (SRSN)}
After denoising process, we propose a light subnetwork for image enlargement and we refer it as Super-Resolution Sub-Network (SRSN). As shown in Figure~\ref{Figure 2}, in order to obtain images with photo-realistic visual quality, a discriminator is attached to form a generative adversarial network. The basic structure of the SRSN is a set of hierarchical residual blocks, which has been widely used in several works~\cite{EDSR,LapSRN,RCAN}. In order to match the dimension, the denoised image is initially up-sampled to the desired dimension by bicubic interpolation.  
\begin{figure}[t]
	\vskip 0.01in
	\begin{center}
		\centerline{\includegraphics[width=0.7\columnwidth]{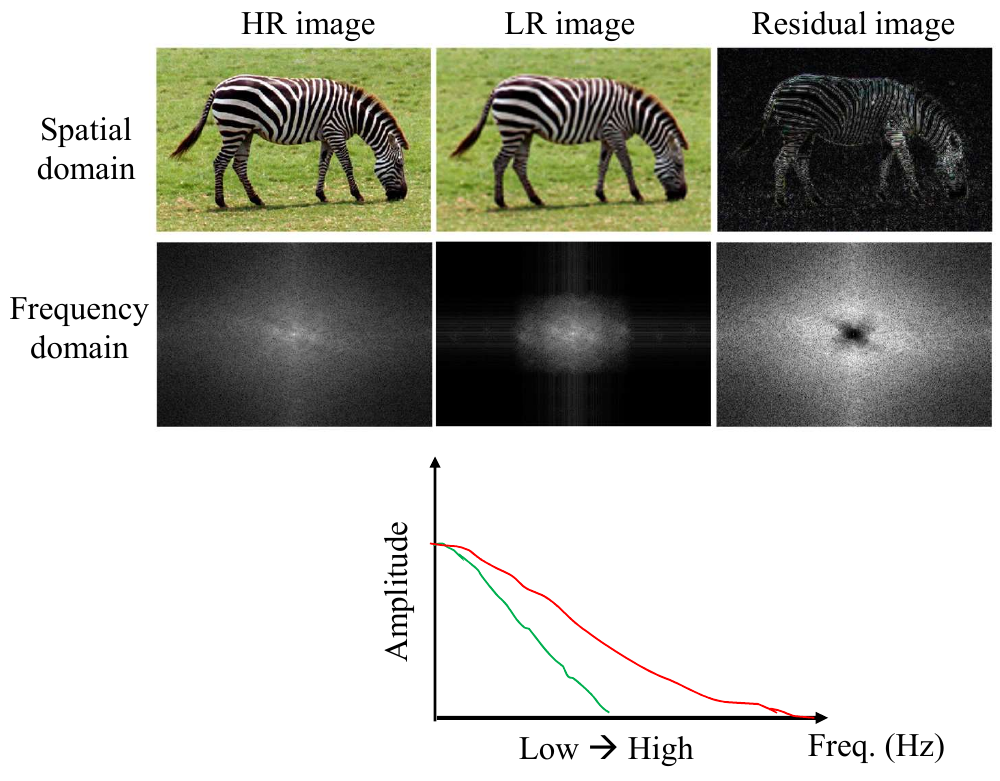}}
		\caption{Spatial and frequency domain differences between HR and LR images.}
		\label{Figure 3}
	\end{center}
	\vskip -0.3in
\end{figure}

Since there is no ground truth HR images to calculate the reconstruction loss (e.g. L1-norm loss), we propose a novel cycle training strategy that comes from the back-projection theory, which is different from the previous related works~\cite{CycleGAN,CycleGAN2017} Let us use Figure~\ref{Figure 3} to interpret the image SR from the signal processing aspect. The HR image contains both low- and high-frequency components. The former represents the basic structure of the image while the latter represents complex patterns, like edges and textures. Assuming that we obtain a ``perfect'' SR image, we down-sample it to generate the corresponding LR image. Relatively, the LR image stands for the low frequency information of the SR image. The super-resolution process can be updated by back projecting the residues learned from the down-sampled SR image and the original LR image. We therefore form a cycle to check the network for its ability of making robust super-resolution. Mathematically, we have the following loss function to demonstrate the cycle training strategy.
\begin{small}
\begin{equation}\tag*{(8)}
\begin{matrix} 
\begin{split}
L_{MAE}=&\sum_{c=1}^C \sum_{h=1}^H \sum_{w=1}^W |s(\mathbf{Y}_{c,h,w})-\mathit{g}(\mathbf{X}_{c,h,w})| \\
& + |\mathbf{Y}_{c,h,w}-\mathbb{Y}_{c,h,w}| \\
& where \  \mathbb{Y}=\mathit{f}(s(\mathbf{Y})), \mathbf{Y}=\mathit{f}(\mathit{g}(\mathbf{X})) 
\end{split}
\end{matrix} 
\label{Equation 8}
\end{equation}
\end{small}\\
where $L_{MAE}$ is the pixel based Mean Absolute Errors (MAE), \textit{f} and \textit{g} are the SRSN and DAE parameters, \textit{C}, \textit{H} and \textit{W} are the size of SR images. \textit{s} is the down-sampling operator with Bicubic process for simplicity. $\mathbf{Y}$ is the output SR image and $\mathbb{Y}$ is the back projected SR image. Equation~\ref{Equation 8} is a loose constraints on image super-resolution because there is no ground truth to compute the actual loss. The first term in Equation~\ref{Equation 8} is to guarantee the low frequency consistency and the second term is to force the back projected SR image to be close to the SR image. We use Bicubic for down-sampling because the real down-sampling operator is too complicated to model it. It is unnecessary to ensure the exact difference between down-sampled SR estimation $\mathit{f}(s(\mathbf{Y}))$ and denoised LR image $\mathit{g}(\mathbf{X})$ because the network is trained to converge until the estimated LR is close to ground truth LR.

On the other hand, Equation~\ref{Equation 8} does not give a strong supervision to the high-frequency reconstruction. It is crucial to give a constraint on the high frequency component. We add a discriminator to take both reference image and the SR image as inputs for real-fake classification. Its objective is to distinguish the high frequency differences between the SR and HR images. Considering that there is no corresponding HR 
images, for $\alpha\times$ image SR, we randomly crop a $\alpha H \times\alpha W$ larger patch from the reference image to match the dimension of the SR result. To encourage the network to pick up photo-realistic features, we also use pre-trained VGG19 to extract the feature map for estimation. Both SR and the denoised LR images are sent to VGG19 to output the feature maps obtained by the 4th convolution layer before the 5th ``Maxpooling'' layer. The SR feature maps are down-sampled by $\alpha\times$ to match the LR feature maps. The total training loss is described as follows,
\begin{small}
\begin{equation}\tag*{(9)}
\begin{matrix} 
\begin{split}
\mathbf{L}=& \lambda \left \| \phi_i(\mathit{f}(\mathit{g}(X)))-s(\phi_i(\mathit{g}(X))) \right \|_1^1 \\
& + \eta log[1-D_{\theta_D} (G_{\theta_G} (\mathit{g}(X)))] + \mathbf{L_{MAE}} 
\end{split}
\end{matrix} 
\label{Equation 9}
\end{equation}
\end{small}

where $\lambda$ and $\eta$ are two weighting parameters to balance the VGG feature loss and adversarial loss. 
$\theta_G$ and $\theta_D$ are the learnable parameters of the generator and discriminator, respectively. $\phi_i$ represents the features from the i-th convolutional layer.

\section{Experiments}
\subsection{Data Preparation and Network Implementation}
We conducted experiments with the training data provided by NTIRE2020 Real World Super-Resolution Challenge~\cite{NTIRE2020RWSRchallenge}. The training dataset is formed by Flickr2K and DIV2K. They both contain images with resolution larger than 1000$\times$1000. The Flickr2K dataset is not only degraded by unknown factors but also down-sampled $4\times$ by an unknown operator. The objective is to learn a mapping function to map from the source domain (Flickr2K) to the target domain (DIV2K). W extracted patches from the training dataset with the size of 128$\times$128. For the discriminator of the proposed SRSN, we extracted 512$\times$512 patches as references for training. For testing, we not only gave focus on super-resolution, but also denoising for real images. The testing datasets include BSD68~\cite{BSD68}, Set5~\cite{Set5}, Urban100~\cite{Urban100}, NTIRE2019 Real Images~\cite{NTIRE2019} and NTIRE2020 validation~\cite{NTIRE2020RWSRchallenge}. Among them, BSD68 is a common dataset for image denoising. Set5 and Urban100 are used for image super-resolution. NTIRE2019 Real Images contains 20 images captured by different cameras with various noise and blurring effects. NTIRE2020 validation includes images with the same degradation as the training images.

To efficiently super-resolve the LR image, we used the pre-trained VGG19 (remove the fully connected layers) as the encoder of the proposed DAE. The length of the latent vector is 512. The decoder is made of 2 deconvolution layers with kernel size 6, stride 4 and padding 1, and 3 residual blocks with kernel size 3, stride 1 and padding 1. The Super-Resolution Sub-Network (SRSN) has 4 residual blocks. Each residual block contains  64 kernels of size 3, stride 1 and padding 1. In the following experiments, we will demonstrate that the proposed dSRVAE can achieve comparable or even better SR performance.

We conducted our experiments using Pytorch 1.4 on a PC with two NVIDIA GTX1080Ti GPUs. During the training, we set the learning rate to 0.0001 for all layers. The batch size was set to 16 for 1$\times10^6$ iterations. For optimization, we used Adam with the momentum to 0.9 and the weight decay of 0.0001. The executive codes and more experimental results can be found in the following link: \url{https://github.com/Holmes-Alan/dSRVAE}. We encourage readers to download the SR results from the link for better visual comparison.

\subsection{Image Denoising}
For our proposed dSRVAE, the Denoising AutoEncoder (DAE) is trained for removing noise from the input LR image. To demonstrate the capability of using Variational AutoEncoder, we tested two different datasets: BSD68 and NTIRE2019. Note that BSD68 is a clean dataset that can be added with additional random noise for evaluation and NTIRE2019 dataset was used for image super-resolution. We used it because the dataset was captured in real life by cameras. It reflects the real image processing scenario so that it can be used for denoising evaluation. In order to evaluate the efficiency of the DAE for denoising, we design another plain convolutional network made of multiple convolutional layers for comparison and we refer it as \textit{net-CNN}. We also experimented on other state-of-the-art image denosing approaches and show the comparison in the following table.
\begin{table}[b]
\caption{Quantitative comparison of different networks for image denoising. {\color{red}Red} indicates the best results.}
\label{Table 1}
\vskip -0.1in
\begin{center}
\begin{small}
\scalebox{0.7}{
\begin{tabular}{ccccccc}
\hline
\multicolumn{7}{c}{BSD68($\sigma=$15)}                                            \\ \hline
Algorithm   & BM3D   & DnCNN  & FFDNet & TNRD  & net-CNN & DAE (ours) \\
PSNR (dB)   & 31.07  & 31.73  & 31.63  & 31.42 & 31.56   & {\color{red}31.81}         \\ \hline
\multicolumn{7}{c}{NTIRE2019}                                            \\ \hline
\multicolumn{3}{c}{Algorithm} & DnCNN  & PD    & net-CNN & DAE (ours)           \\
\multicolumn{3}{c}{PSNR (dB)}  & 29.30  & 29.53 & 29.36   & {\color{red}29.54}        \\ \hline
\end{tabular}
}
\end{small}
\end{center}
\vskip -0.3in
\end{table}

\begin{figure*}[h]
	\vskip 0.01in
	\begin{center}
		\centerline{\includegraphics[width=0.7\textwidth]{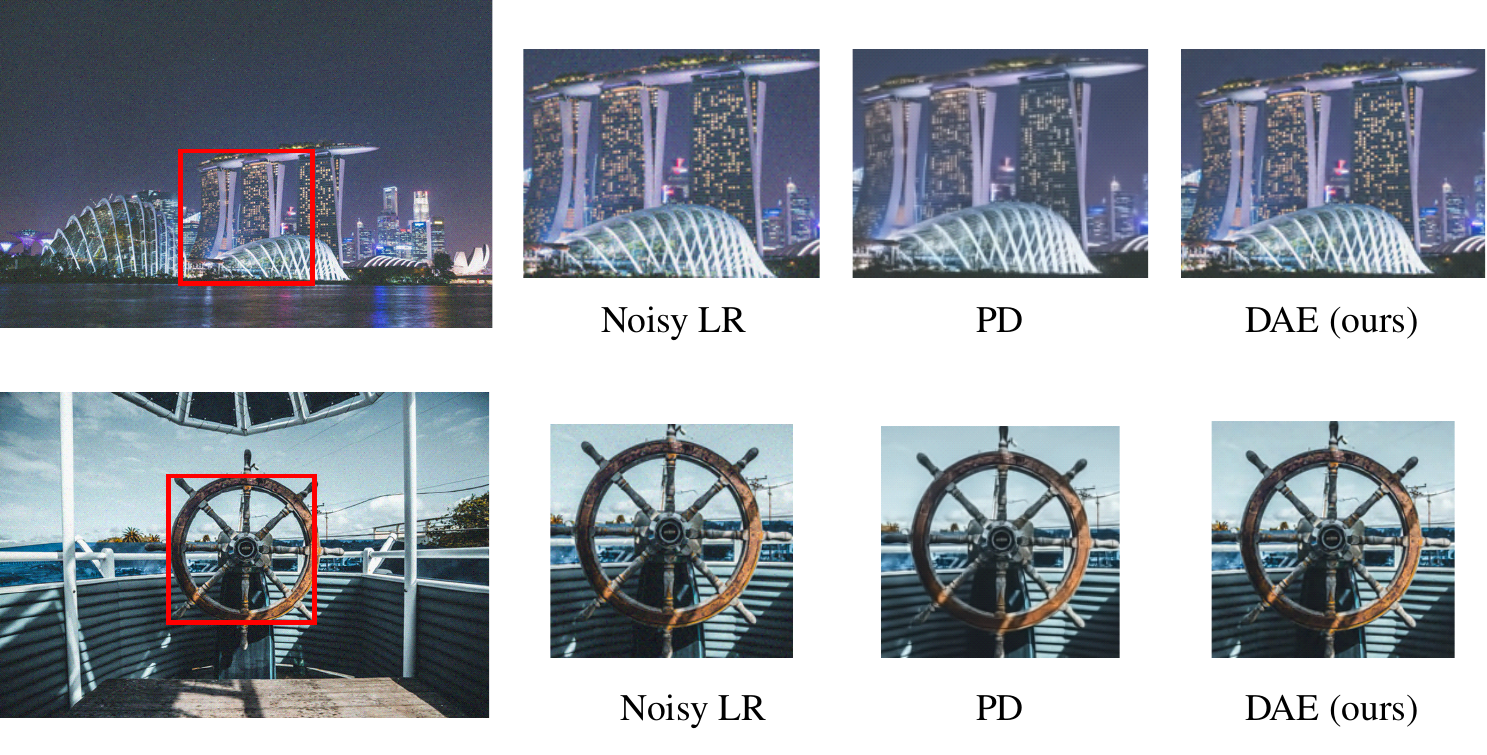}}
		\caption{Visualization of image denoising on NTIRE2020 validation images. Enlarged red boxes are included for better comparison.}
		\label{Figure 4}
	\end{center}
	\vskip -0.3in
\end{figure*}

In Table~\ref{Table 1}, we compare our approach with five classic denoising approaches with BSD68 and NTIRE2019. From the PSNR results, it shows that the proposed DAE achieves better performance. Note that we tested the BSD68 with Gaussian noise of variance 15. We did not test using other Gaussian noise levels because our objective is not for additive noise removal. Our target is to illustrate the denoising capability of our proposed DAE model. In order to show the denoising ability on real image with unknown noise, we tested the NTIRE2020 validation dataset and show the visual comparison in Figure~\ref{Figure 4}.

It can be seen from Figure~\ref{Figure 4} that both approaches can remove the noise in the background, like the sky in these two images. More interestingly, using proposed DAE can preserve as much details as possible while the PD approach tends to oversmooth the edgy areas (check the windows on the buildings and the textures on the wheel) to remove the noise. 

\subsection{Image Super-Resolution}
More importantly, to prove the effectiveness of the proposed dSRVAE network, we conducted experiments by comparing some of the state-of-the-art SR algorithms: Bicubic, SRGAN~\cite{SRGAN}, ESRGAN~\cite{ESRGAN} and BlindSR~\cite{BlindSR}. PSNR and SSIM were used to evaluate the quantitative distortion performance and PI score~\cite{PIRM} was used to indicate the perception performance. Generally, PSNR and SSIM were calculated by converting the RGB image to YUV and taking the Y-channel image for estimation. PI takes the RGB image for estimation. We only focus on 4$\times$ image SR. All approaches were reimplemented using the codes provided by the corresponding authors.

In the following sections, we will give evaluation on different down-sampling scenarios, including ideal bicubic down-sampling, camera simulation and unknown degradation.

\begin{figure*}[h]
	\vskip 0.01in
	\begin{center}
		\centerline{\includegraphics[width=0.7\textwidth]{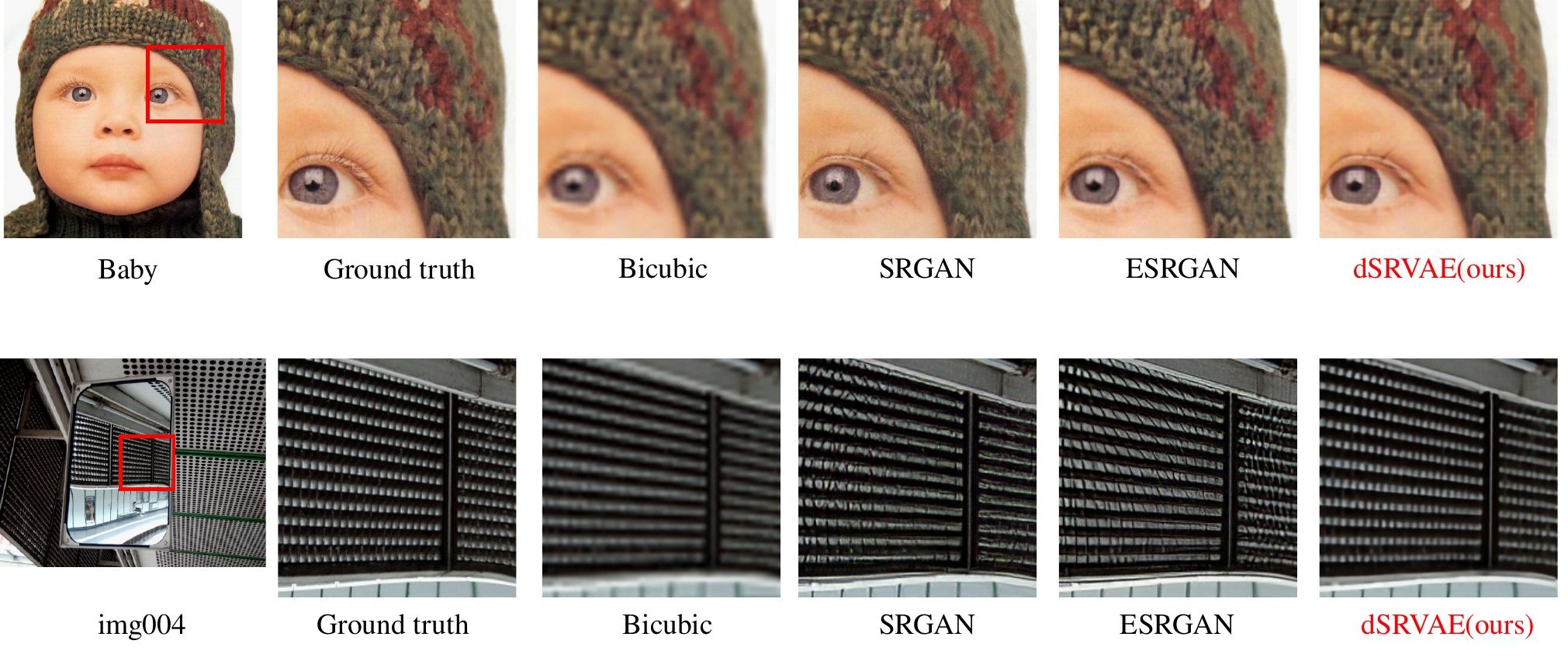}}
		\caption{Visualization of 4$\times$ image super-resolution on Set5 and Urban100 images. Enlarged red boxes are included for better comparison.}
		\label{Figure 5}
	\end{center}
	\vskip -0.3in
\end{figure*}
\noindent\textbf{Analysis on ideal Bicubic down-sampled SR}\\
First, for classic image SR, we assume bicubic as a standard down-sampling operator for image SR. With sufficient training images and deeper structures, a lot of works have been proposed to improve SR performance. Initially, our network was trained in unsupervised way for real image. It cannot be used to compare most of the existing SR approaches. For a fair comparison, we modified our network to take paired LR and HR images for supervised training. The MAE loss function in Equation~\ref{Equation 8} was modified to calculate the errors between SR and HR. Adversarial loss was also used for photo-realistic image SR. For the sake of objective measurement, Table~\ref{Table 2} shows the quantitative results among different approaches.

\begin{table}[t]
	\caption{Quantitative comparison of different networks for 4$\times$ image super-resolution on Set5 and Urban100. {\color{red}Red} indicates the best results.}
	\label{Table 2}
	\vskip -0.1in
	\begin{center}
		\begin{small}
			\scalebox{0.9}{
				\begin{tabular}{cccccc}
					\hline
					\multicolumn{2}{c}{\multirow{2}{*}{Algorithm}} & \multicolumn{2}{c}{Set5} & \multicolumn{2}{c}{Urban100} \\
					\multicolumn{2}{c}{}                           & PSNR        & PI         & PSNR          & PI           \\ \hline
					Bicubic             & \multirow{4}{*}{4$\times$}       & 28.42       & 7.370      & 23.64         & 6.944        \\
					ESRGAN              &                          & 30.47       & 3.755      & 24.36         & {\color{red}3.484}       \\
					SRGAN               &                          & 29.40       & {\color{red}3.355}     & 24.41         & 3.771        \\
					dSRGAN(ours)        &                          & {\color{red}31.46}       & 4.836      & {\color{red}26.33}         & 4.481        \\ \hline
				\end{tabular}
			}
		\end{small}
	\end{center}
	\vskip -0.3in
\end{table}

\begin{figure*}[h]
	\vskip 0.01in
	\begin{center}
		\centerline{\includegraphics[width=0.75\textwidth]{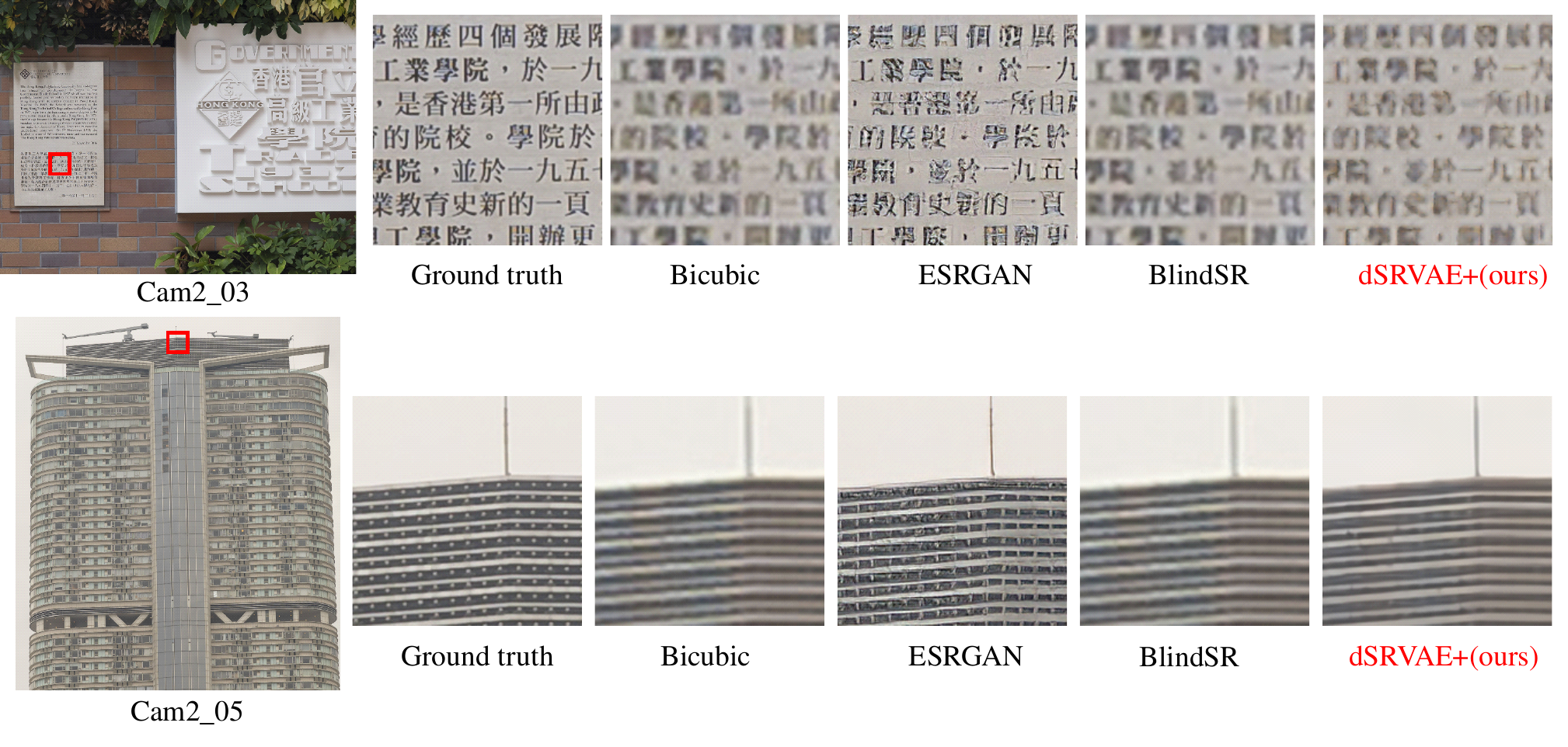}}
		\caption{Visualization of 4$\times$ image super-resolution on NTIRE2019 validation. Enlarged red boxes are included for better comparison.}
		\label{Figure 6}
	\end{center}
	\vskip -0.3in
\end{figure*}

Table~\ref{Table 2} lists the PSNR and PI score on Set5 and Urban100 for 4$\times$ SR. Higher PSNR means lower distortion and lower PI score means better visual quality. The results show that the proposed network can achieve comparable performance to state-of-the-art image SR approaches. Since all approaches focus on perceptual quality, we use Figure~\ref{Figure 5} to demonstrate the visualization comparison.
Figure~\ref{Figure 5} shows two examples from Set5 and Urban100. We can see that using the proposed dSRVAE can provide photo-realistic details, like the textures on the hat of the \textit{Baby} and the metal bars in the mirror of \textit{img004}. 

\noindent\textbf{Analysis on real image captured by cameras}\\
In this part, we will show visual comparison on NTIRE2019 dataset. It is a dataset that contains images captured by different cameras under different conditions. We use this dataset to test the generalization of our proposed dSRVAE. Our comparison includes supervised approaches (ESRGAN, SRGAN) and BlindSR, a novel blind image SR approach trained on different blur and down-sampling kernels. 

From the results in Figure~\ref{Figure 6}, we can see that the proposed dSRVAE not only can effectively remove the noise from the LR image, but also preserves the original pattern without severe distortion. For example, ESRGAN can generate much sharper edges on the texts of image \textit{Cam2\_03} but with some bizarre patterns. On the other hand, compared with BlindSR, dSRVAE can provide sharper reconstruction without distorting the pattern of the texts. Similar results can also be observed for image \textit{cam2\_05}. 
\noindent\textbf{Analysis on real image with unknown degradation factors}\\
Finally, let us make a comparison on NTIRE2020 testing images. This dataset contains 100 high-resolution images. The LR images were down-sampled 4$\times$ by unknown degradation factors, including noise and artifacts. It is more complicated than simple bicubic down-sampling or camera simulation scenarios. Without knowing the ground truth images, we provide visualization of different SR approaches to illustrate the SR performance.
\begin{figure*}[h]
	\vskip 0.01in
	\begin{center}
		\centerline{\includegraphics[width=0.85\textwidth]{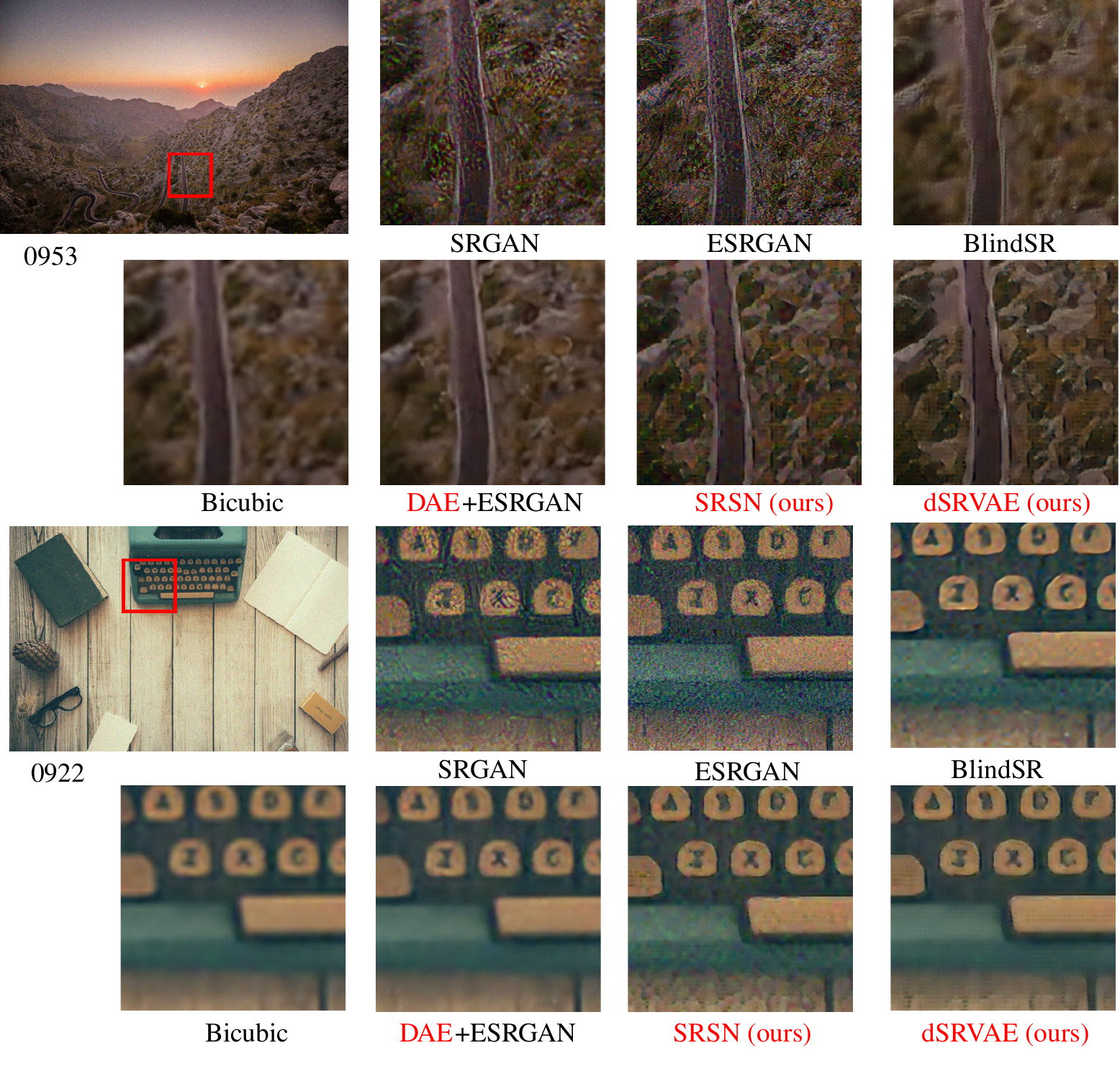}}
		\caption{Visualization of 4$\times$ image super-resolution on NTIRE2020 validation. Enlarged red boxes are included for better comparison.}
		\label{Figure 7}
	\end{center}
	\vskip -0.3in
\end{figure*}
Based on our assumption, joint learn denoising and super-resolution can be beneficial for real image SR because we always encounter various noise on the real image that cannot just be resolved by a single SR model, especially when the noise is signal-dependent. It is useful to disentangle the correlation between noise and signal for other processes. To demonstrate the performance of our proposed dSRVAE, we compare with two perceptual image SR approaches (ESRGAN and SRGAN) and one blind image SR approach (BlindSR). Our target is to test whether the proposed ``first-denosing-then-SR'' strategy works. In Figure~\ref{Figure 7}, dSRVAE is referred to our final result. We separate the Denoising AutoEncoder and Super-Resolution Sub-Network to independently test whether they work. We refer SRSN as the one withouting using DAE for denoising and ``DAE+ESRGAN'' as the one first using DAE for denoising and then use ESRGAN for SR.

Figure~\ref{Figure 7} shows the results on images ``0922'' and ``0953''. We can see that the keyboard and the road are much better reconstructed by dSRVAE. SRGAN and ESRGAN were trained on using clean LR images down-sampled by bicubic so that they cannot handle the noise. BlindSR, on the other hand, can partially remove some noise and can not provide much improvement on the textures. DAE+ESRGAN helps to remove the noise with a little blurring effect because it was not trained end-to-end. Using SRSN only would be affected by the noise. Our approach, dSRVAE, can effectively remove the noise, and also improve the overall quality. 

\section{Discussion}
In this paper, we propose an unsupervised real image super-resolution approach with Generative Variational AutoEncoder. Two key points were introduced: 1) Variational AutoEncoder for image denoising and, 2) cycle training strategy for unsupervised image super-resolution. In order to obtain photo-realistic SR images, we combine variational autoencoder and generative adversarial network for joint image denoising and super-resolution. Experimental results show that our proposed real image denoising and Super-Resolution via Variational AutoEncoder (dSRVAE) approach achieves good perceptual performance on different datasets. More importantly, results on testing NTIRE2019 and NTIRE2020 datasets show that the proposed dSRVAE can handle real image super-resolution for practical applications.

{\small
\bibliographystyle{ieee_fullname}
\bibliography{paper_review}

\begin{thebibliography}{10}\itemsep=-1pt

\bibitem{ZSSR}
Michal~Irani Assaf~Shocher, Nadav~Cohen.
\newblock "zero-shot" super-resolution using deep internal learning.
\newblock In {\em The IEEE Conference on Computer Vision and Pattern
  Recognition (CVPR)}, June 2018.

\bibitem{Set5}
Marco Bevilacqua, Aline Roumy, Christine Guillemot, and Marie line
  Alberi~Morel.
\newblock Low-complexity single-image super-resolution based on nonnegative
  neighbor embedding.
\newblock In {\em Proceedings of the British Machine Vision Conference}, pages
  135.1--135.10. BMVA Press, 2012.

\bibitem{PIRM}
Yochai Blau, Roey Mechrez, Radu Timofte, Tomer Michaeli, and Lihi
  Zelnik{-}Manor.
\newblock 2018 {PIRM} challenge on perceptual image super-resolution.
\newblock {\em CoRR}, abs/1809.07517, 2018.

\bibitem{NTIRE2019}
Jianrui Cai, Shuhang Gu, Radu Timofte, and Lei Zhang.
\newblock Ntire 2019 challenge on real image super-resolution: Methods and
  results.
\newblock In {\em The IEEE Conference on Computer Vision and Pattern
  Recognition (CVPR) Workshops}, June 2019.

\bibitem{BlindSR}
Victor Cornill{\`e}re, Abdelaziz Djelouah, Wang Yifan, Olga Sorkine-Hornung,
  and Christopher Schroers.
\newblock Blind image super resolution with spatially variant degradations.
\newblock {\em ACM Transactions on Graphics (proceedings of ACM SIGGRAPH
  ASIA)}, 38(6), 2019.

\bibitem{SRCNN}
C. {Dong}, C.~C. {Loy}, K. {He}, and X. {Tang}.
\newblock Image super-resolution using deep convolutional networks.
\newblock {\em IEEE Transactions on Pattern Analysis and Machine Intelligence},
  38(2):295--307, Feb 2016.

\bibitem{NCSR}
W. {Dong}, L. {Zhang}, G. {Shi}, and X. {Li}.
\newblock Nonlocally centralized sparse representation for image restoration.
\newblock {\em IEEE Transactions on Image Processing}, 22(4):1620--1630, April
  2013.

\bibitem{KSVD}
M. {Elad} and M. {Aharon}.
\newblock Image denoising via sparse and redundant representations over learned
  dictionaries.
\newblock {\em IEEE Transactions on Image Processing}, 15(12):3736--3745, Dec
  2006.

\bibitem{FrequencySF}
Manuel Fritsche, Shuhang Gu, and Radu Timofte.
\newblock Frequency separation for real-world super-resolution.
\newblock {\em 2019 IEEE/CVF International Conference on Computer Vision
  Workshop (ICCVW)}, pages 3599--3608, 2019.

\bibitem{DBPN}
Muhammad Haris, Greg Shakhnarovich, and Norimichi Ukita.
\newblock Deep back-projection networks for super-resolution.
\newblock {\em CoRR}, abs/1803.02735, 2018.

\bibitem{Urban100}
J. {Huang}, A. {Singh}, and N. {Ahuja}.
\newblock Single image super-resolution from transformed self-exemplars.
\newblock In {\em 2015 IEEE Conference on Computer Vision and Pattern
  Recognition (CVPR)}, pages 5197--5206, June 2015.

\bibitem{VDSR}
Jiwon Kim, Jung~Kwon Lee, and Kyoung~Mu Lee.
\newblock Accurate image super-resolution using very deep convolutional
  networks.
\newblock {\em CoRR}, abs/1511.04587, 2015.

\bibitem{VAE}
Diederik~P. Kingma, Tim Salimans, and Max Welling.
\newblock Variational dropout and the local reparameterization trick.
\newblock {\em ArXiv}, abs/1506.02557, 2015.

\bibitem{LapSRN}
Wei{-}Sheng Lai, Jia{-}Bin Huang, Narendra Ahuja, and Ming{-}Hsuan Yang.
\newblock Deep laplacian pyramid networks for fast and accurate
  super-resolution.
\newblock {\em CoRR}, abs/1704.03915, 2017.

\bibitem{SRGAN}
Christian Ledig, Lucas Theis, Ferenc Huszar, Jose Caballero, Andrew~P. Aitken,
  Alykhan Tejani, Johannes Totz, Zehan Wang, and Wenzhe Shi.
\newblock Photo-realistic single image super-resolution using a generative
  adversarial network.
\newblock {\em CoRR}, abs/1609.04802, 2016.

\bibitem{EDSR}
Bee Lim, Sanghyun Son, Heewon Kim, Seungjun Nah, and Kyoung~Mu Lee.
\newblock Enhanced deep residual networks for single image super-resolution.
\newblock {\em CoRR}, abs/1707.02921, 2017.

\bibitem{SRRMF}
Z. {Liu}, W. {Siu}, and Y. {Chan}.
\newblock Fast image super-resolution via randomized multi-split forests.
\newblock In {\em 2017 IEEE International Symposium on Circuits and Systems
  (ISCAS)}, pages 1--4, May 2017.

\bibitem{ISCAS17}
Z. Liu, W. Siu, and Y. Chan.
\newblock Fast image super-resolution via randomized multi-split forests.
\newblock In {\em 2017 IEEE International Symposium on Circuits and Systems
  (ISCAS)}, pages 1--4, May 2017.

\bibitem{RefSR}
Z. {Liu}, W. {Siu}, and Y. {Chan}.
\newblock Reference based face super-resolution.
\newblock {\em IEEE Access}, 7:129112--129126, 2019.

\bibitem{ABPN}
Zhi{-}Song Liu, Li{-}Wen Wang, Chu{-}Tak Li, and Wan{-}Chi Siu.
\newblock Attention based back projection network for image super-resolution.
\newblock {\em 2019 IEEE Conference on Computer Vision Workshops (ICCVW2019)},
  2019.

\bibitem{HBPN}
Zhi{-}Song Liu, Li{-}Wen Wang, Chu{-}Tak Li, and Wan{-}Chi Siu.
\newblock Hierarchical back projection network for image super-resolution.
\newblock {\em 2019 IEEE Conference on Computer Vision and Pattern Recognition
  Workshops (CVPRW2019)}, abs/1906.06874, 2019.

\bibitem{lugmayrICCVW2019}
Andreas Lugmayr, Martin Danelljan, and Radu Timofte.
\newblock Unsupervised learning for real-world super-resolution.
\newblock In {\em ICCV Workshops}, 2019.

\bibitem{AIM2019RWSRchallenge}
Andreas Lugmayr, Martin Danelljan, Radu Timofte, et~al.
\newblock Aim 2019 challenge on real-world image super-resolution: Methods and
  results.
\newblock In {\em ICCV Workshops}, 2019.

\bibitem{NTIRE2020RWSRchallenge}
Andreas Lugmayr, Martin Danelljan, Radu Timofte, et~al.
\newblock Ntire 2020 challenge on real-world image super-resolution: Methods
  and results.
\newblock {\em CVPR Workshops}, 2020.

\bibitem{BSD68}
D. Martin, C. Fowlkes, D. Tal, and J. Malik.
\newblock A database of human segmented natural images and its application to
  evaluating segmentation algorithms and measuring ecological statistics.
\newblock In {\em Proc. 8th Int'l Conf. Computer Vision}, volume~2, pages
  416--423, July 2001.

\bibitem{VGG}
Karen Simonyan and Andrew Zisserman.
\newblock Very deep convolutional networks for large-scale image recognition.
\newblock In {\em International Conference on Learning Representations}, 2015.

\bibitem{A+}
Radu Timofte, Vincent~De Smet, and Luc~Van Gool.
\newblock {A+:} adjusted anchored neighborhood regression for fast
  super-resolution.
\newblock In Daniel Cremers, Ian~D. Reid, Hideo Saito, and Ming{-}Hsuan Yang,
  editors, {\em Computer Vision - {ACCV} 2014 - 12th Asian Conference on
  Computer Vision, Singapore, Singapore, November 1-5, 2014, Revised Selected
  Papers, Part {IV}}, volume 9006 of {\em Lecture Notes in Computer Science},
  pages 111--126. Springer, 2014.

\bibitem{ESRGAN}
Xintao Wang, Ke Yu, Shixiang Wu, Jinjin Gu, Yihao Liu, Chao Dong, Chen~Change
  Loy, Yu Qiao, and Xiaoou Tang.
\newblock {ESRGAN:} enhanced super-resolution generative adversarial networks.
\newblock {\em CoRR}, abs/1809.00219, 2018.

\bibitem{ss}
Yunxuan Wei, Shuhang Gu, Yawei Li, and Longcun Jin.
\newblock Unsupervised real-world image super resolution via domain-distance
  aware training.
\newblock In {\em 2020 IEEE Computer Society Conference on Computer Vision and
  Pattern Recognition (CVPR)}, 2020.

\bibitem{CycleGAN}
Yuan Yuan, Siyuan Liu, Jiawei Zhang, Yongbing Zhang, Chao Dong, and Liang Lin.
\newblock Unsupervised image super-resolution using cycle-in-cycle generative
  adversarial networks.
\newblock {\em CoRR}, abs/1809.00437, 2018.

\bibitem{Denoise_SR}
Kai Zhang, Wangmeng Zuo, and Lei Zhang.
\newblock Learning a single convolutional super-resolution network for multiple
  degradations.
\newblock {\em CoRR}, abs/1712.06116, 2017.

\bibitem{LPIPS}
Richard Zhang, Phillip Isola, Alexei~A Efros, Eli Shechtman, and Oliver Wang.
\newblock The unreasonable effectiveness of deep features as a perceptual
  metric.
\newblock In {\em CVPR}, 2018.

\bibitem{Demosaic_SR_2}
Xuaner~Cecilia Zhang, Qifeng Chen, Ren Ng, and Vladlen Koltun.
\newblock Zoom to learn, learn to zoom.
\newblock {\em CoRR}, abs/1905.05169, 2019.

\bibitem{RCAN}
Yulun Zhang, Kunpeng Li, Kai Li, Lichen Wang, Bineng Zhong, and Yun Fu.
\newblock Image super-resolution using very deep residual channel attention
  networks.
\newblock {\em CoRR}, abs/1807.02758, 2018.

\bibitem{CRFSR}
L. {Zhi-Song} and W. {Siu}.
\newblock Cascaded random forests for fast image super-resolution.
\newblock In {\em 2018 25th IEEE International Conference on Image Processing
  (ICIP)}, pages 2531--2535, Oct 2018.

\bibitem{ICIP18}
L. {Zhi-Song} and W. {Siu}.
\newblock Cascaded random forests for fast image super-resolution.
\newblock In {\em 2018 25th IEEE International Conference on Image Processing
  (ICIP)}, pages 2531--2535, 2018.

\bibitem{Demosaic_SR_1}
Ruofan Zhou, Radhakrishna Achanta, and Sabine S{\"{u}}sstrunk.
\newblock Deep residual network for joint demosaicing and super-resolution.
\newblock {\em CoRR}, abs/1802.06573, 2018.

\bibitem{CycleGAN2017}
Jun-Yan Zhu, Taesung Park, Phillip Isola, and Alexei~A Efros.
\newblock Unpaired image-to-image translation using cycle-consistent
  adversarial networks.
\newblock In {\em Computer Vision (ICCV), 2017 IEEE International Conference
  on}, 2017.

\end{thebibliography}
}

\end{document}